\documentclass[letterpaper, 10 pt, conference]{ieeeconf}  

\IEEEoverridecommandlockouts                              

\usepackage{cite}
\usepackage{graphicx} 
\usepackage{amsmath} 
\usepackage{cite}
\usepackage{graphicx}
\usepackage{amsmath, amssymb}
\usepackage{booktabs}
\usepackage{multirow}
\usepackage{multicol}
\usepackage{hyperref}
\usepackage{booktabs}

\usepackage{caption}
\usepackage{gensymb}
\usepackage{amsmath}
\usepackage{xcolor}

\title{\LARGE \bf
Learning a Geometric Representation for Data-Efficient \\ Depth Estimation via Gradient Field and Contrastive Loss
}

\author{Dongseok Shim and H. Jin Kim$^{*}$

\thanks{This research was supported by Unmanned Vehicles Core Technology Research and Development Program through the National Research  Foundation of Korea(NRF) and Unmanned Vehicle Advanced Research Center(UVARC) funded by the Ministry of Science and ICT, the Republic of Korea(NRF-2020M3C1C1A01086411)}%
\thanks{Authors are with the Department
of Mechanical and Aerospace Engineering, Seoul National University,
Gwanak-gu, Seoul, 08826, Korea. E-mail: 
        {\tt\small \{tlaehdtjr01, hjinkim\}@snu.ac.kr} ($^{*}$Corresponding author)}%
}

\begin{document}

\maketitle
\begin{abstract}

Estimating a depth map from a single RGB image has been investigated widely for localization, mapping, and 3-dimensional object detection.
 Recent studies on a single-view depth estimation are mostly based on deep Convolutional neural Networks (ConvNets) which require a large amount of training data paired with densely annotated labels. 
 Depth annotation tasks are both expensive and inefficient, so it is inevitable to leverage RGB images which can be collected very easily to boost the performance of ConvNets without depth labels.
 However, most self-supervised learning algorithms are focused on capturing the semantic information of images to improve the performance in classification or object detection, not in depth estimation.
 In this paper, we show that existing self-supervised methods do not perform well on depth estimation and propose a gradient-based self-supervised learning algorithm with momentum contrastive loss to help ConvNets extract the geometric information with unlabeled images. As a result, the network can estimate the depth map accurately with a relatively small amount of annotated data.
 To show that our method is independent of the model structure, we evaluate our method with two different monocular depth estimation algorithms.
 Our method outperforms the previous state-of-the-art self-supervised learning algorithms and shows the efficiency of labeled data in triple compared to random initialization on the NYU Depth v2 dataset.

\end{abstract}

\section{Introduction}
Depth estimation is an essential element for mobile robots such as a drone or autonomous driving car to navigate paths, deviate obstacles, and understand environments by scene.
 Especially, weight and cost issues of high-quality depth sensors like LiDAR or RGB-D camera motivate the research on depth estimation using a monocular camera.

\begin{figure}
    \centering
    \includegraphics[width=0.5\textwidth]{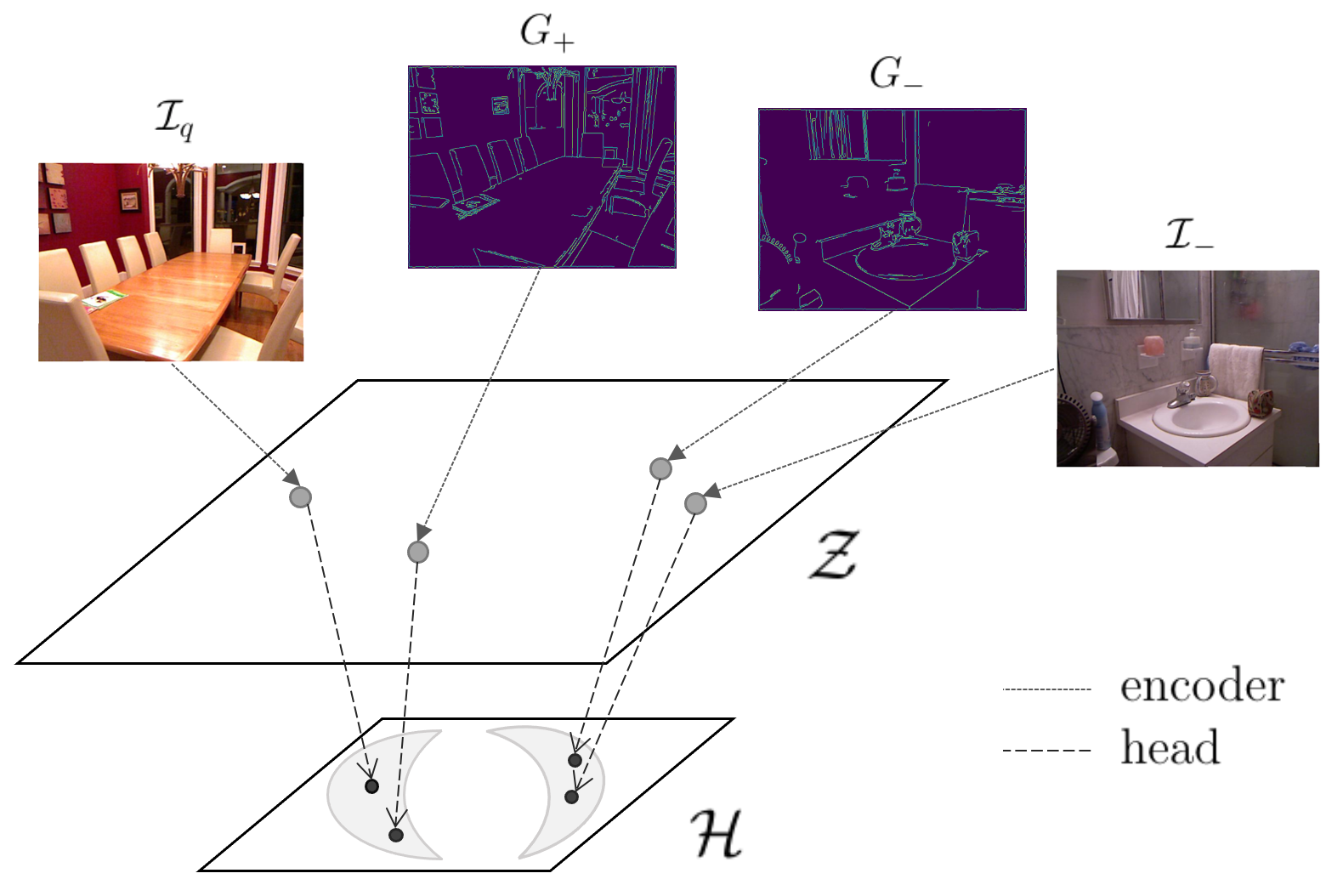}
    \caption{Our proposed method trains an encoder to learn a geometric visual representation of an image $\mathcal{I}$ with its gradient field $G$. The encoder maps the image and its gradient field to the feature space $\mathcal{Z}$ and the head projects its feature $z$ to the low-dimensional head space $\mathcal{H}$  to prevent overfitting. By contrastive loss, the representation of a query image $\mathcal{I}_{q}$ in the head space $\mathcal{H}$ becomes closer to the representation of the positive pair $G_{+}$ and farther to the representation of the negative pair $G_{-}$ from the other image $\mathcal{I}_{-}$.}
    \label{fig:thumnail}
\end{figure}

In recent years, depth estimation from a single RGB image has been influenced by the dramatic success of the deep convolutional neural networks\cite{lecun1998gradient}.
 A complex depth estimation network for higher accuracy\cite{fu2018deep,alhashim2018high,chang2019deep} and an efficient network for a real-time or on-device operation\cite{laina2016deeper,wofk2019fastdepth} have been both achieved.
 However, training a depth estimation network requires a large number of images with pixel-wise depth annotated labels, which costs very intense and expensive efforts to collect.

Due to such difficulties, a need for leveraging unlabeled RGB images is increasing to boost the performance of the ConvNet with a small amount of labeled data.
 Previous self-supervised learning algorithms such as rotation\cite{gidaris2018unsupervised}, exemplar\cite{dosovitskiy2015discriminative} pretext tasks or contrastive learning based approach\cite{he2020momentum,chen2020simple} are mainly focused on extracting semantic features from the image to increase the performance on classification, object detection, or semantic segmentation. Self-supervised learning methods have also been applied for depth estimation\cite{zhou2017stereo, garg2016unsupervised, godard2019digging, zhou2017unsupervised}, but they all require stereo left-right paired images\cite{zhou2017stereo, garg2016unsupervised} or images from the monocular camera which are consecutively collected so that a large portion of images overlap\cite{ godard2019digging, zhou2017unsupervised}.
 
Therefore, we propose a self-supervised learning method for ConvNet to learn geometric features of an image with unlabeled independent RGB images for increasing the performance in depth estimation.
 We utilize the Sobel kernel and Canny edge binary mask \cite{canny1986computational} to generate gradient fields of the image as the positive and negative pairs for momentum contrast learning\cite{he2020momentum}. 
 To do so, the ConvNet learns the relationship between the RGB image and its gradient field by distinguishing the source image of the gradient field from the other RGB images. 
 The proposed method is general and model-agnostic in that it is compatible with any parametric-model-based depth estimation network regardless of the structure of encoder and decoder. 
 Our method also outperforms existing state-of-the-art self-supervised visual representation learning in the monocular depth estimation and achieves a better performance than random initialization by \cite{he2015delving} with 3 times fewer annotated depth labels. To the best of our knowledge, it is the first approach to pre-train the depth estimation network with independently sampled images in an unsupervised manner.

\begin{figure*}[h]
\includegraphics[width=1.0\textwidth]{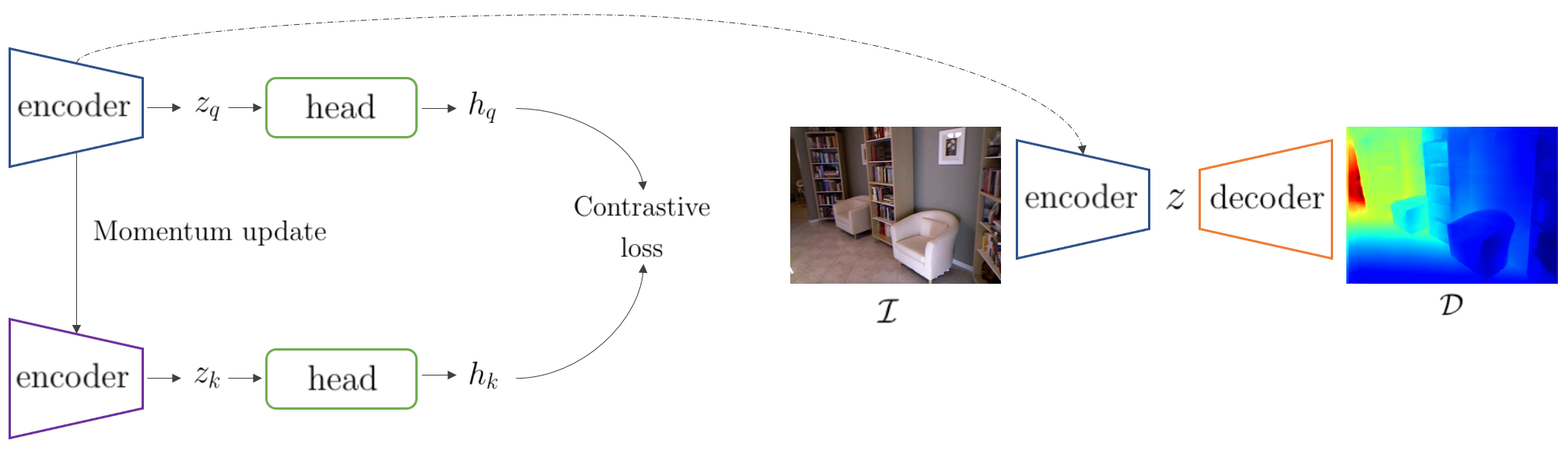}
\caption{\textbf{Momentum contrastive learning} Two encoders with the same structure are adopted for contrastive learning. A query encoder is updated by the back-propagation with contrastive loss and the key encoder is updated by momentum. The query encoder is used as the feature extractor of the depth estimation network after contrastive pre-training for geometric representation learning.}
\label{fig:gradfield}
\end{figure*}

In short, our contribution can be summarized as follows:
\begin{itemize}
\item We show that the previous self-supervised visual representation learning with independent images do not perform well on the depth estimation task.
\item The proposed gradient-based momentum contrastive learning enables the ConvNet to capture the geometric representation of the image regardless of its model structure.
\item Our method outperforms existing self-supervised learning based pre-training method both in accuracy and error metric of monocular depth estimation and achieves three times better data-efficiency compared to random initialization.
\end{itemize}

We evaluate our method on NYU Depth v2 dataset\cite{Silberman:ECCV12} with two different monocular depth estimation networks to prove that our method is model-agnostic and it is applicable to various networks from the high-quality complex networks to the real-time efficient networks. The implementation of the paper in PyTorch and its pre-trained models are available at \texttt{\href{https://github.com/dsshim0125/grmc}{https://github.com/dsshim0125/grmc}}.


\section{Related Work}

In this section, we summarize the past research on monocular depth estimation and recap visual representation learning algorithms which are categorized into self-supervised pretext and contrastive learning.

\subsection{Monocular Depth Estimation}
Recent studies on monocular depth estimation use deep learning algorithms with a large-scale labeled dataset. Due to the properties of the ConvNet, the size of the receptive field decreases as the image passes multiple layers with different sizes of the convolution kernels and strides. Therefore, most depth estimation networks leverage an encoder-decoder structure to increase the size of the output depth map.

The two-scale ConvNets which generate coarse and fine depth maps each to feed additional information by concatenation from coarse to fine networks have been suggested in \cite{eigen2014depth}. Later, their work is extended to three-scale ConvNets for other auxiliary estimation tasks like normal prediction or  segmentation\cite{eigen2015predicting}. In \cite{laina2016deeper}, a deep residual network ResNet\cite{he2016deep} is used as the encoder to extract features from the images and has used a novel up-projection algorithm for efficient and faster training. \cite{fu2018deep} formulates the depth estimation problem as the quantized ordinal regression and \cite{alhashim2018high} applies a U-Net\cite{ronneberger2015u} with DenseNet\cite{huang2017densely} as the encoder to generate a high-quality depth map with complex neural networks. 

Self-supervised learning based depth estimation has also been studied by estimating a camera pose or disparity maps \cite{ zhou2017unsupervised,zhou2017stereo, garg2016unsupervised, godard2019digging}, but they need paired stereo images\cite{zhou2017stereo, garg2016unsupervised, godard2017unsupervised} or at least sequences of the monocular images\cite{ zhou2017unsupervised, godard2019digging}. It means that the algorithms above cannot be trained with independently sampled monocular images.

\subsection{Self-Supervised Visual Representation Learning}
As most images are unlabeled, significant research efforts aim to train neural networks more efficiently with a small number of labeled dataset. Self-supervised learning formulates a pretext task using only unlabeled data so that the network can learn useful visual representation from the image before task-specific supervised learning such as classification or object detection.

In \cite{doersch2015unsupervised}, a single image is divided into several non-overlapped patches and the networks are trained to predict their relative positions. Follow-up studies \cite{noroozi2016unsupervised, noroozi2018boosting} also generate patches but the patches are randomly shuffled and the pretext task is to recover the original images by predicting the permutation of the patches which is similar to solving jigsaw puzzles.

A rotation pretext that rotates the input image is introduced in \cite{gidaris2018unsupervised}, and the network predicts the degree of applied rotation. The rotation degree is an element of a set that has a finite cardinality of specific numbers, so this poses the rotation pretext task as a classification problem, not a regression.

An exemplar task has been proposed in \cite{dosovitskiy2015discriminative} which decreases the distance between the original RGB image (seed image) and its heavily augmented image and increases the distance between the seed and augmented image from the other RGB images. It allows the network to extract a visual representation which is invariant to a wide range of image transformations.

\subsection{Contrastive Learning}
Recently, a visual representation learning algorithm based contrastive learning has achieved huge success in the image classification task. 
Contrastive learning \cite{hadsell2006dimensionality} is an approach to learn representations by enforcing the attractive force to positive pairs and the repulsive forces to negative pairs.
The proposed method in \cite{henaff2020data} divides the image into several non-overlapping patches and the task is to predict the pixel values of the next unseen patch. They use the target patch as the positive pair and patches from the other location in the same image or patches from the other images as negative pairs for the contrastive learning.
\cite{he2020momentum} and  \cite{chen2020simple}
impose multiple image transformations on RGB images and set the positive pair as the augmented image from the same image and the negative pair as the augmented image from the other image. The difference is that \cite{he2020momentum} uses 2 encoders with the same structure but \cite{chen2020simple} only uses a single encoder.


\section{Method}
In this section, we introduce the learning algorithm of the momentum-based contrastive learning with two encoders by \cite{he2020momentum}, compare it with our proposed method, and present the gradient field with the modified Canny edge detector\cite{canny1986computational} which is used as the positive and negative pairs for contrastive learning.

\subsection{Momentum Contrastive Learning}

We pre-train the encoder of the depth estimation network with the contrastive loss to learn the representation of the image without any supervisory signals. We adopt momentum contrastive learning suggested in MoCo\cite{he2020momentum}, with 2 encoders which are query and key encoders respectively.

The encoder maps the positive and negative pairs of the image to the feature space $\mathcal{Z}$ to extract the latent vector $z$ which contains the compressed information of the input data. 
Although the latent vector $z$ has a low dimension, it is still vulnerable to overfitting to a specific training data domain, so the $head$ module is adopted which is a 2-layer fully-connected network followed by ReLU non-linear activation function\cite{chen2020simple}. 
The head module projects the latent vector $z$ to the feature space $\mathcal{H}$ which has much lower dimensionality with similar representation compared to the original feature space $\mathcal{Z}$. 
The contrastive loss is then calculated by the similarity between the projected latent vectors $h$ of positive and negative pairs in the low-dimensional feature space.

MoCo\cite{he2020momentum} leverages a dynamic dictionary as a queue of latent vector $h$ of input data $K = \{h_{k_{0}}, h_{k_{1}}, h_{k_{2}}, ...\}$ so that the contrastive learning can be less dependent on the batch size. 
The encoded key value $h_{k}$ is not discarded after training but stacked in the dictionary to formulate large negative pairs during training. 
The size of the dynamic dictionary $K$ is fixed, so the most outdated values are removed when the number of the latent vectors in the dictionary exceeds the maximum size. 
To avoid a rapid change of the key value $h_{k}$ from the same RGB image $\mathcal{I}$ for stable representation learning,  MoCo\cite{he2020momentum} separates two encoders each for the query and key data. 
The key encoder is not updated by contrastive loss but rather by the momentum update as to keep the key latent vector $h_{k}$ consistent.

\begin{figure}[t]
    \centering
    \includegraphics[width=0.5\textwidth]{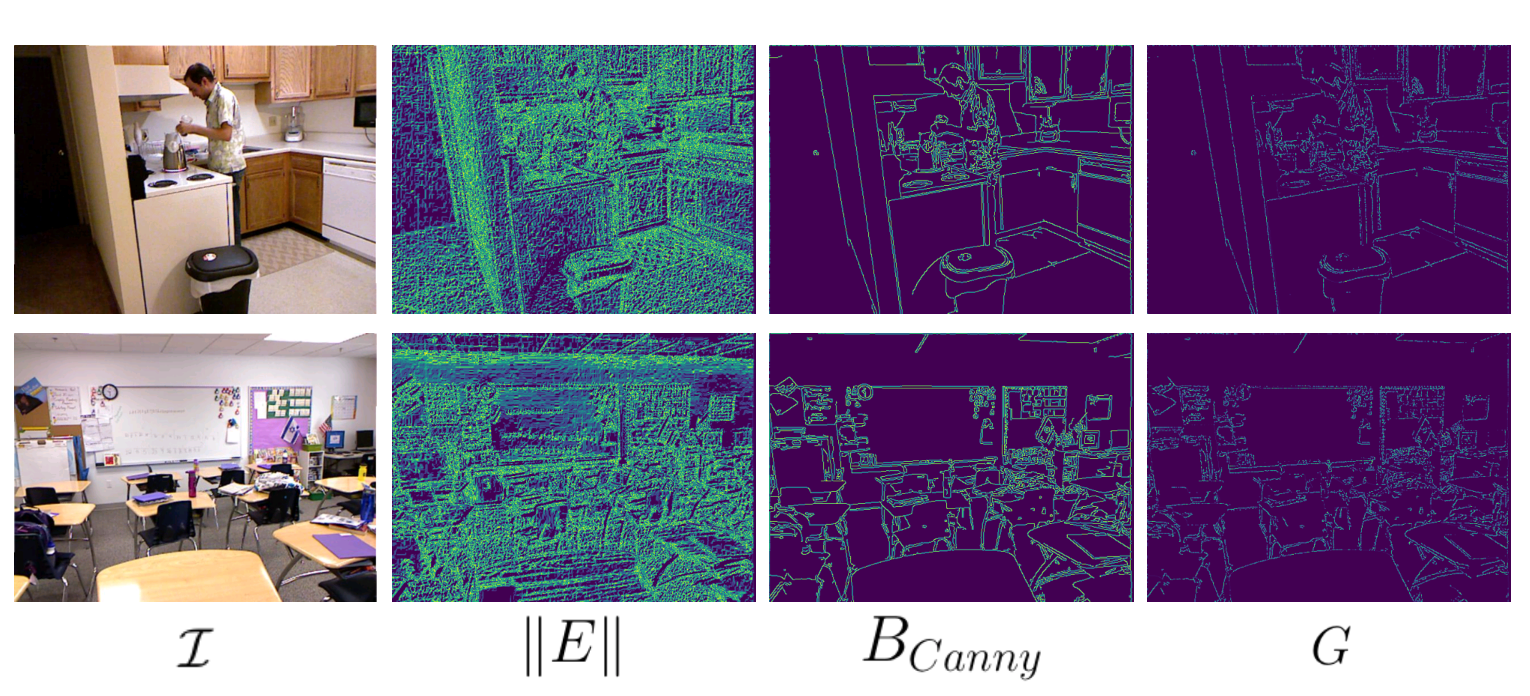}
    \caption{\textbf{Gradient field generation} We generate a gradient field $G$ of the image $\mathcal{I}$ for contrastive learning by a pixel-wise multiplication of binary masks $B_{Canny}$ and the magnitude of the gradient $\|E\|$, which are the result and the byproduct of Canny edge detection algorithm respectively.}
    \label{fig:gradient}
\end{figure}

\begin{figure*}[t]
\centering
\includegraphics[width=1.0\textwidth]{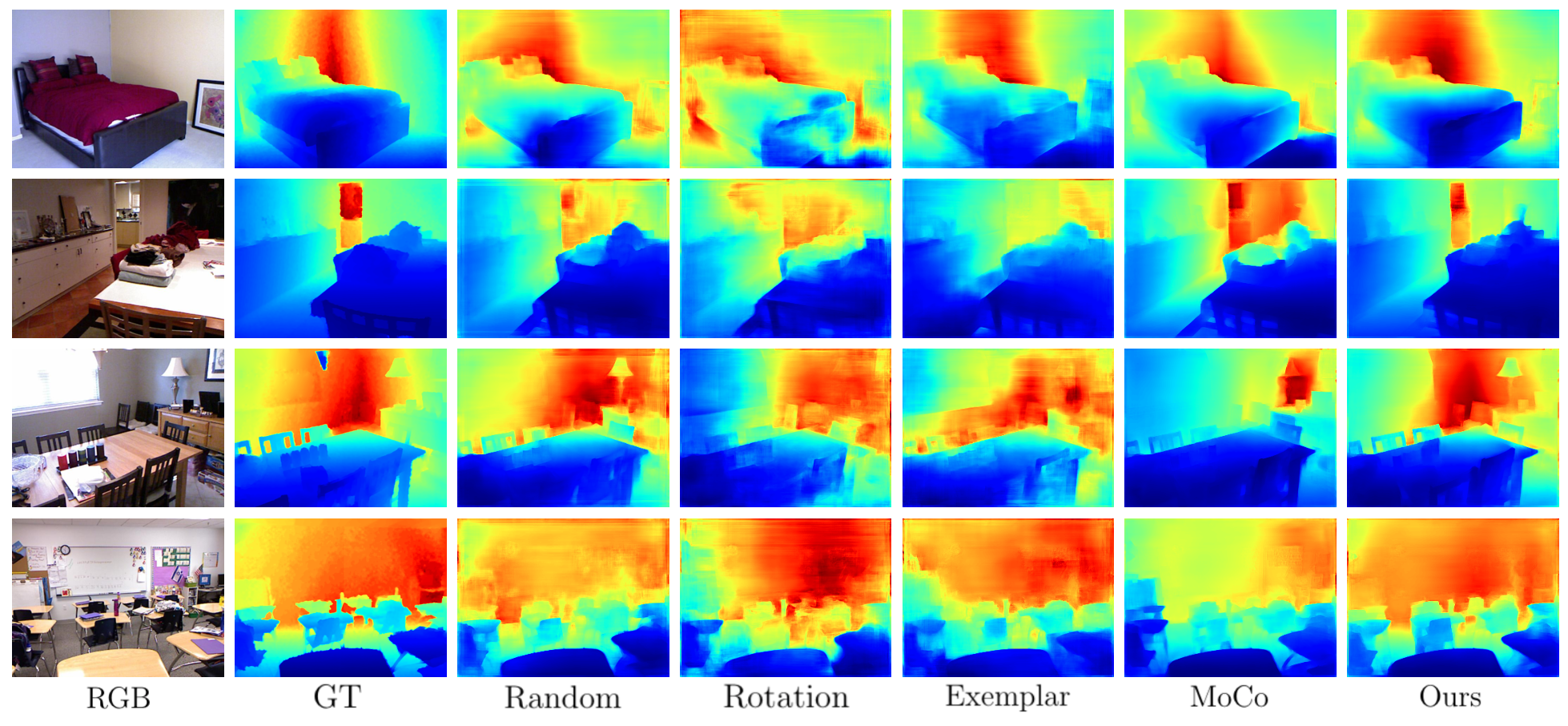}
\caption{\textbf{Visualization of depth estimation on NYU Depth v2 dataset} Qualitative results of monocular depth estimation pre-trained by Rotation\cite{gidaris2018unsupervised}, Exemplar\cite{dosovitskiy2015discriminative}, MoCo\cite{he2020momentum}, and our method compared to the random initialization\cite{he2015delving} and the ground truth depth map.}
\label{fig:depth}
\end{figure*}

As MoCo\cite{he2020momentum} measures the similarity between the query latent vector $h_{q}$ and the key latent vector $h_{k}$ with a dot product, the contrastive loss function $\mathcal{L}_{q}$ can be formulated 
as Eq. (\ref{eqn:infonce}) which is called  InfoNCE\cite{oord2018representation}. The subscript $+$ indicates the positive pair, and $\tau$ is a temperature parameter to control the concentration of the distribution\cite{wu2018unsupervised}.

\begin{equation}
    \mathcal{L}_{q} = -\mathrm{log} \frac{\mathrm{exp}(h_{q}\cdot h_{k_{+}}/\tau)}{ \Sigma\mathrm{exp}^{K}_{i=0}(h_{q} \cdot h_{k_{i}}/\tau)}
    \label{eqn:infonce}
\end{equation}

The difference between our method and MoCo\cite{he2020momentum} is that we choose the query data as the RGB image $\mathcal{I}_{q}$ and the key data as the gradient field $G$ whereas MoCo\cite{he2020momentum} uses the augmented images by color distortion both as the query and key data.
By adopting the gradient field as the key data for momentum contrastive learning, the encoder can learn the geometric representation of the image rather than semantic representation. We distinguish the type of query data from the type of key data, which are RGB image and gradient field respectively.
It is because we intend to update the query encoder by the back-propagation\cite{ rumelhart1986learning} only with the error signal from RGB image that is the input of our final objective of the network, i.e., monocular depth estimation. After training with the momentum contrastive loss, the query encoder is used as the initial point of depth estimation network as shown in Fig. \ref{fig:gradfield}.

\subsection{Gradient Field}

Previous self-supervised learning algorithms are mainly focused on capturing the semantic information of the image, so most of them formulate the unlabeled pretext task by imposing wide range of image transformations like HSV-space color randomization\cite{dosovitskiy2015discriminative}, Inception cropping\cite{szegedy2015going} or Gaussian blur\cite{chen2020simple}. In this paper, we exploit a gradient field of the image as positive and negative pairs of contrastive learning so that the network can extract the visual geometric features.

To generate the gradient field $G$ of the image $\mathcal{I}$, we use Canny edge detector\cite{canny1986computational} to extract the edge of the image.
Unlike the standard Canny algorithm which generates a binary mask to filter the magnitude of the gradient from the Sobel operator by comparing its value with the neighbor pixels, we modify the Canny detector to extract the value of the magnitude of the dominant gradient itself as well as its location so that the pixel of the gradient field has different intensity value according to its edge dominance. The procedure of generating a gradient field can be mathematically expressed as Eq. (\ref{eq1}):
\begin{equation}\label{eq1}
  \begin{gathered}
    \mathcal{I} \in \mathbb{R}^{h\times w},\: \mathcal{I}_{u}, \, \mathcal{I}_{v} \in \mathbb{R}^{h\times w},        \\
    \|E\| = \sqrt{\mathcal{I}_{u}^{2} + \mathcal{I}_{v}^{2}},  \\
    G = B_{Canny}\otimes \, \|E\|,\\
  \end{gathered}
\end{equation}
where 
\begin{align*}
    \mathcal{I}_{u} = \frac{\partial \mathcal{I}}{\partial u},\; \mathcal{I}_{v} = \frac{\partial \mathcal{I}}{\partial v}.
\end{align*}
$\|E\|$ and $B_{Canny}$ denote the magnitude of the gradient from the Sobel operator and the binary result of the Canny algorithm respectively, and the operator $\otimes : (\mathbb{R}^{h\times w}, \mathbb{R}^{h\times w}) \mapsto \mathbb{R}^{h\times w}$ indicates the pixel-wise multiplication. Lastly, as the range of the magnitude of the gradient field is different from the input RGB image, we both normalize pixel values of the image and gradient field from 0 to 1 for a faster and stabler model learning. We visualize the magnitude of the gradient, binary mask from Canny detector and our proposed gradient field in Fig. \ref{fig:gradient}.

In short, we pre-train the encoder of the depth estimation network with the momentum-based contrastive learning by\cite{he2020momentum} and the gradient field from the modified Canny algorithm\cite{canny1986computational} to learn the geometric representation of the images. Our proposed momentum contrastive learning with the gradient field is summarized in Fig. \ref{fig:thumnail}.


 \begin{table*}[t]
    \centering
    \begin{tabular}{cccccccc}
    \midrule[1.0pt]
    Model&Method&\multicolumn{3}{c}{Accuracy metric}&\multicolumn{3}{c}{Error metric}\\
    \cmidrule(lr){3-5}\cmidrule(lr){6-8}
    \multirow{9}{*}{Alhashim \textit{et al.}\cite{alhashim2018high}}& & $\delta<1.25$ & $\delta<1.25^{2}$ & $\delta<1.25^{3}$ &rel&rms&$\mathrm{log}_{10}$\\
    \midrule[0.5pt]
    &Random\cite{he2015delving} & 0.743&0.932&0.980&0.175&0.599&0.073\\
    &ImageNet$^{*}$\cite{ILSVRC15} & 0.826 & 0.967 & 0.999 & 0.131&0.484&0.057\\
    \cmidrule[0.03pt]{2-8}
    &Rotation\cite{gidaris2018unsupervised} & 0.693&0.911&0.974&0.189&0.666&0.082\\
    &Exemplar\cite{dosovitskiy2015discriminative} &0.516&0.810&0.938&0.255&0.923&0.117\\
    &MoCo\cite{he2020momentum} & 0.755 & 0.933 & 0.981 & 0.168 & 0.592 & 0.071\\
    &Ours & \textbf{0.801} & \textbf{0.952} & \textbf{0.986} & \textbf{0.147} & \textbf{0.532} & \textbf{0.062}\\
    \midrule[0.5pt]
    \multirow{7}{*}{Laina \textit{et al.}\cite{laina2016deeper}}
    &Random\cite{he2015delving} & 0.672&0.899&0.969&0.207&0.710&0.087\\
    &ImageNet$^{*}$\cite{ILSVRC15} & 0.773 & 0.950 & 0.989 & 0.157&0.553&0.067\\
    \cmidrule[0.03pt]{2-8}
    &Rotation\cite{gidaris2018unsupervised} & 0.608& 0.871& 0.961& 0.239& 0.791&0.098 \\
    &Exemplar\cite{dosovitskiy2015discriminative} &0.598&0.861&0.954&0.242&0.822&0.101\\
    &MoCo\cite{he2020momentum} & 0.692 & 0.909 & 0.974 & 0.201 & 0.668 & 0.082\\
    &Ours & \textbf{0.709} & \textbf{0.917} & \textbf{0.976} & \textbf{0.192} & \textbf{0.656} & \textbf{0.080}\\
    \midrule[1.0pt]
    \end{tabular}
    
\caption{\textbf{Comparison to prior self-supervised learning algorithms} The evaluation has been done on the official test split of NYU Depth v2\cite{Silberman:ECCV12}. * indicates supervised pre-training with ground truth labels and all the other methods are pre-trained in an unsupervised manner. The accuracy and error metrics can be defined as $\delta$: $\mathrm{max}(\frac{\hat{y}_{p}}{y_{p}}, \frac{y_{p}}{\hat{y}_{p}})$; rel: $\frac{1}{n}\Sigma^{n}_{p}\frac{\|y_{p} - \hat{y}_{p}\|}{y}$; rms: $\sqrt{\frac{1}{n}\Sigma^{n}_{p}(y_{p} - \hat{y}_{p})^{2}}$; log$_{10}$: $\frac{1}{n}\Sigma^{n}_{p}\|\mathrm{log}_{10}(y_{p}) - \mathrm{log}_{10}(\hat{y}_{p})\|$; where $y_{p}$, $\hat{y}_{p}$ are pixels of depth map $y$ and its estimation $\hat{y}$. All the results are re-experimented under the same condition for a fair evaluation.}
\label{tab:table1}
\end{table*}

\section{Experiments}
To demonstrate the effectiveness of our self-supervised pre-training, we first explain the dataset and implementation details, and present multiple experiments to compare our method with existing self-supervised algorithms.

\subsection{Dataset}
We both train the encoder in an unsupervised manner and fine-tune the entire depth estimation network with supervisory signals on the NYU Depth v2\cite{Silberman:ECCV12} dataset. The dataset consists of indoor 120K train samples and 654 test samples collected with Microsoft Kinect. We train our method on the 50K subsets of the train sample and evaluate the method on the entire 654 test samples. A resolution of the dataset is 640$\times$480 and we do not resize the input RGB image, but the output depth map is down-sampled to half resolution 320$\times$240 for training speed and computational efficiency. During the evaluation, we trim the test input image with a pre-defined center cropping by \cite{eigen2014depth} for a precise evaluation.

\subsection{Implementation Details}
We implement our self-supervised pre-training and monocular depth estimation networks\cite{alhashim2018high, laina2016deeper} on a public deep learning platform PyTorch\cite{paszke2019pytorch}.
We re-experimented both the self-supervised and depth estimation tasks to eliminate the factors that can affect the performance of the network except for the pre-trained weights of the encoder.
For training an encoder with momentum-based contrastive learning, we set the batch size as 64, the size of the dynamic dictionary $K$ as 16384, and the temperature parameter $\tau$ as 0.07. We adopt the SGD optimizer with the learning rate of 0.015, the momentum of 0.9, and set the weight decay as 0.0001.
For fine-tuning the depth estimation network, we train depth estimation networks with the same input size 640$\times$480 and use the Adam optimizer with the learning rate of 0.0001. We set the batch size as 4 for training and set the batch size as 1 for evaluation.

\begin{table}[t]
\setlength\tabcolsep{1pt}
\centering
\begin{tabular}{cc} 
\centering
\includegraphics[width=0.50\linewidth]{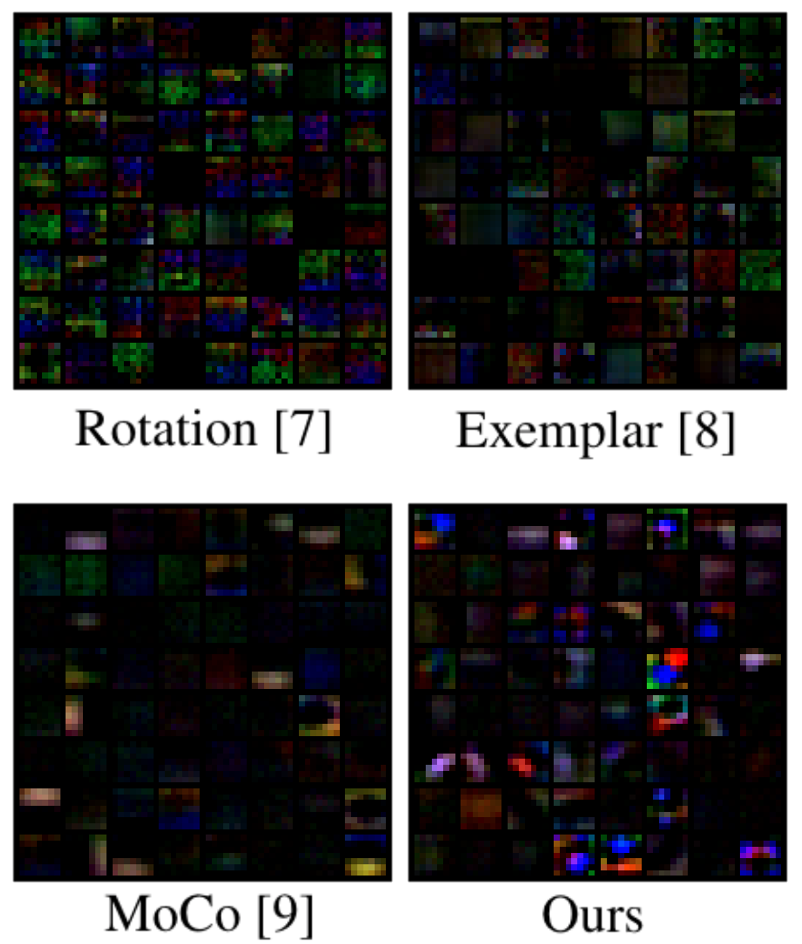} &
\includegraphics[width=0.50\linewidth]{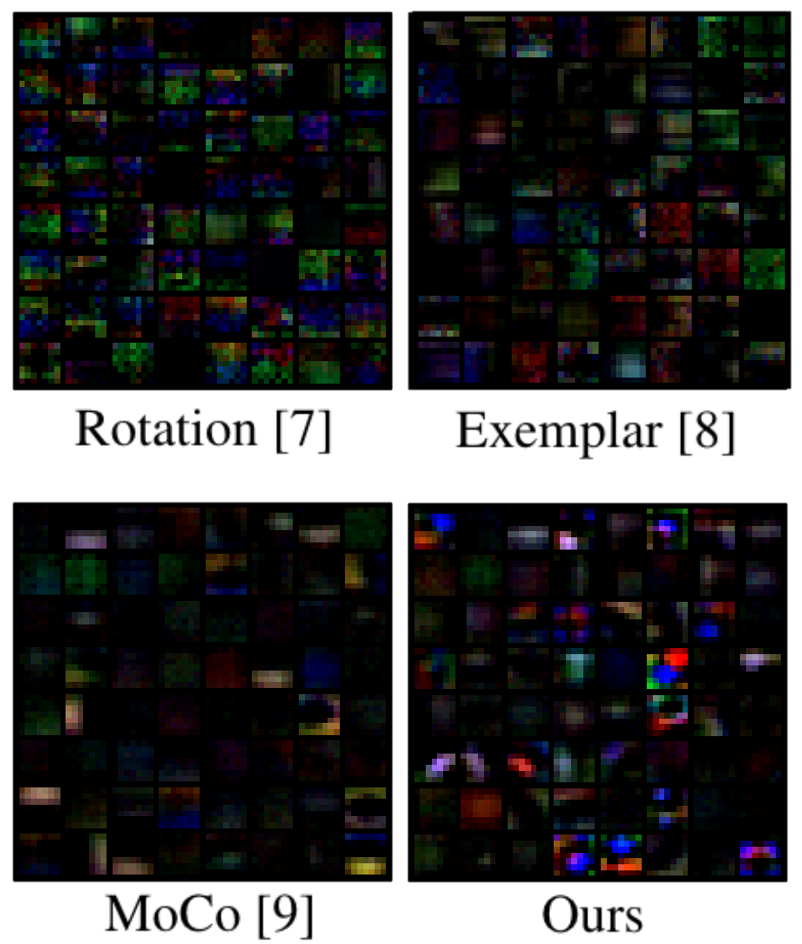} \\
(a) Self-supervised & (b) Fine-tune

\end{tabular}
\captionof{figure}{\textbf{Kernel visualization of the encoder} First 64 convolution kernels learned by DenseNet-161\cite{huang2017densely} trained on (a) the self-supervised pretext for encoder pre-training and (b) the depth estimation fine-tuning.}
\label{fig:kernel}
\end{table}

\subsection{Performance Evaluations}
We compare the results on monocular depth estimation according to the pre-trained weights of the encoder with two different structures of networks,  \cite{alhashim2018high} \cite{laina2016deeper}. 
The work in \cite{alhashim2018high} adopts DenseNet-161\cite{huang2017densely} as the encoder for high-quality complex depth estimation and \cite{laina2016deeper} uses ResNet-50\cite{he2016deep} as the encoder and its up-projection module as the decoder for a real-time operation.
TABLE \ref{tab:table1} shows some meaningful results.
Our method seems to show worse performance than \textit{supervised} ImageNet\cite{ILSVRC15} pre-training due to the difference in the  dataset size and the presence of the labels: NYU Depth v2\cite{Silberman:ECCV12} contains 120K images whereas ImageNet contains 14M images with classification labels.
However, our method outperforms the existing state-of-the-art self-supervised algorithms on all the quantitative metrics. 
We improves the accuracy by 5\% $\sim$ 30\% trained with \cite{alhashim2018high} and by 2\% $\sim$ 11\% with \cite{laina2016deeper}. 
Interestingly, Rotation\cite{gidaris2018unsupervised} and Exemplar\cite{dosovitskiy2015discriminative} pretext degrade the performance of the depth estimation network compared to random initialization by \cite{he2015delving}.
It indicates that pre-training the network to capture the semantic information of the image for classification or object detection does not always provide a better initialization for other tasks such as depth estimation.
Some qualitative results of our proposed method and prior self-supervised methods trained with \cite{alhashim2018high} are shown in Fig. \ref{fig:depth}.

\subsection{Impact of encoder pre-training}

In Fig. \ref{fig:kernel}, we visualize the kernels of the first layer of DenseNet-161\cite{huang2017densely} learned by our proposed self-supervised pre-training and fine-tuning for depth estimation. We observe that the patterns of the kernels which play a critical role in capturing specific information of the image barely change. The absolute value of the kernels are slightly changed by task adaptation, but the general patterns and shapes of the kernels are not affected by the loss signals from the depth estimation. It indicates that how we pre-train the encoder of the entire network determines the dominant representation of the images that the network can capture regardless of task-specific fine-tuning.

\subsection{Labeled Data-Efficiency}
We evaluate our model on a subset of the entire dataset with annotated depth labels compared to random initialization by \cite{he2015delving}. 
We intentionally restrict the amount of the labeled data varying from 1\% to 10\% of NYU Depth v2\cite{Silberman:ECCV12} for fine-tuning the depth estimation networks. Both training and evaluation have been done with \cite{alhashim2018high} where DenseNet-161\cite{huang2017densely} is used as the encoder. 
From Table \ref{tab:efficient}, our method trained with only 1\% and 5\% of labels outperforms the random initialization by \cite{he2015delving} trained with 3\% and 10\% of labeled data respectively in all the accuracy metrics. 
The gain in data-efficiency of the network increases as the number of the labeled data becomes smaller, where the data- efficiency is triple with 1\% of labeled data and double with 5\% of labeled data. 
It is because the network is trained by the gradient descent, so the parameter initialization becomes more important as the number of labeled training data becomes smaller.

\begin{table}[h]
    \centering
    \begin{tabular}{cccccc}
    \midrule[1.0pt]
    Labeled Data&Metric&1\%&3\%&5\%&10\%\\
    \hline
    Random\cite{he2015delving}&\multirow{2}{*}{$\delta_{1}$}&0.469 & 0.530& 0.623&0.685\\  
    Ours& &0.598& 0.654&0.701& 0.737\\
    \hline
    Random\cite{he2015delving}&\multirow{2}{*}{$\delta_{2}$}& 0.767& {0.813}& 0.878&{0.909}\\  
    Ours& &{0.870}&0.904 & {0.921}&0.934\\
    \hline
    Random\cite{he2015delving}& \multirow{2}{*}{$\delta_{3}$}&0.909& {0.931}& 0.960&{0.973}\\  
    Ours& &{0.957}& 0.974& {0.978}&0.984\\
    \hline
    Data-Efficiency&&\multicolumn{2}{c}{$\times 3$}&\multicolumn{2}{c}{$\times 2$}\\
    \midrule[1.0pt]
    
    \end{tabular}
    \caption{\textbf{Data-efficient depth estimation} The accuracy metrics are denoted as $\delta_{i}: \delta <1.25^{i}$.}
    \label{tab:efficient}
\end{table}

\subsection{Evaluation on Domain Generalization}
To demonstrate the domain generalization capability, we evaluate our proposed method on the outdoor dataset even though the depth estimation network is pre-trained and fine-tuned only on the indoor dataset. 
We use Make3D dataset \cite{saxena2008make3d} for evaluation which consists of 534 outdoor images, 400 training data and 134 test data. For evaluation on Make3D, we exploit three commonly used error metrics \cite{karsch2014depth, liu2015learning} and central crop\cite{godard2017unsupervised}. We adopt the depth estimation network from \cite{alhashim2018high} with the encoder of DenseNet-161\cite{huang2017densely} training on NYU Depth v2\cite{Silberman:ECCV12} and no further fine-tuning on images and labels from Make3D is done.

\begin{figure}[t]
    \centering
    \includegraphics[width=0.5\textwidth]{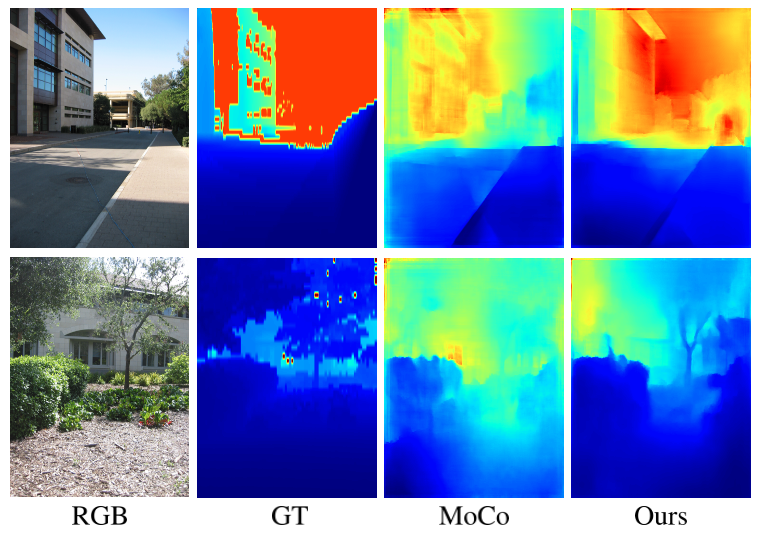}
    \caption{\textbf{Depth prediction on Make3D}. RGB image, ground truth, na\"ive momentum contrast\cite{he2020momentum} and our proposed method. Images and labels from Make3D\cite{saxena2008make3d} dataset are only leveraged for evaluation, not for any training or fine-tuning.}
    \label{fig:make3d}
\end{figure}

Table \ref{tab:make3d} confirms that the proposed pre-training method on Make3D still shows a better performance than the random\cite{he2015delving}, rotation\cite{gidaris2018unsupervised} and MoCo\cite{he2020momentum} initialization. As shown in Fig. \ref{fig:make3d}, depth estimation network pre-trained with our method generates much plausible depth output and preserves more details such as edges of the buildings and branches of the trees compared to random initialization and even to the noisy ground truth depth map.
It is because our proposed method trains ConvNet to extract geometric features of the image such as vertical or horizontal structural information which are robust regardless of indoor and outdoor environments.

\begin{table}[t]
    \centering
    \begin{tabular}{c|ccc}
        \midrule[1.0pt]
         Method&rel & rms & $\mathrm{log}_{10}$   \\
         \hline
         Random\cite{he2015delving}&0.355&10.82&0.479\\
         Rotation\cite{gidaris2018unsupervised} & 0.379  & 11.08   & 0.497      \\
         MoCo\cite{he2020momentum} & 0.349& 10.83&0.476 \\
         Ours & \textbf{0.339}& \textbf{10.68}&\textbf{0.468} \\
         \midrule[1.0pt]
    \end{tabular}
    \caption{\textbf{Performance on Make3D.} Quantitative results on 134 test images of Make3D\cite{saxena2008make3d} dataset. Pixels above 70m in depth are masked out.}
    \label{tab:make3d}
\end{table}


\section{CONCLUSIONS}
 In this paper, we propose a self-supervised pre-training algorithm with momentum-based contrastive learning and gradient field generated from the modified Canny edge detector so that the network can learn the geometric representation of the image for depth estimation. Our method outperforms existing self-supervised learning algorithms both in accuracy and error metrics and enables threefold improvements in data-efficiency. We also show generalization capability to the unseen data from different environment, indoor and outdoor environment. Our proposed method can be further applied to any parametric model which requires a geometric feature such as normal estimation or 3D reconstruction.

\bibliographystyle{IEEEtran}
\bibliography{root.bib}

\end{document}